%% file: ms.tex
\documentclass[letterpaper, 10 pt, conference]{ieeeconf}
\IEEEoverridecommandlockouts    
\input{stachnisslab-latex}

\input{stachnisslab-math}

\usepackage{url}
\usepackage{subfig}

\title{\LARGE \bf A Novel Decomposed Feature-Oriented Framework for Open-Set \\ Semantic Segmentation on LiDAR Data}

\author{Wenbang Deng \and Xieyuanli Chen$^*$ \and Qinghua Yu \and Yunze He \and Junhao Xiao \and Huimin Lu$^*$
  \thanks{W. Deng, X. Chen, Q. Yu, J. Xiao and H. Lu are with the College of Intelligence Science and Technology, National University of Defense Technology, China. Y. He is with Hunan University.}
  \thanks{$^*$X. Chen and H. Lu are corresponding authors.
  }
  \thanks{This work was partly supported by the National Science Foundation of China under Grant 62403478, Young Elite Scientists Sponsorship Program by CAST (No. 2023QNRC001), and Major Project of Natural Science Foundation of Hunan Province under Grant 2021JC0004.
  }%
}

\begin{document}
\maketitle
\thispagestyle{empty}
\pagestyle{empty}

\begin{abstract}
Semantic segmentation is a key technique that enables mobile robots to understand and navigate surrounding environments autonomously. However, most existing works focus on segmenting known objects, overlooking the identification of unknown classes, which is common in real-world applications. In this paper, we propose a feature-oriented framework for open-set semantic segmentation on LiDAR data, capable of identifying unknown objects while retaining the ability to classify known ones. We design a decomposed dual-decoder network to simultaneously perform closed-set semantic segmentation and generate distinctive features for unknown objects. The network is trained with multi-objective loss functions to capture the characteristics of known and unknown objects. Using the extracted features, we introduce an anomaly detection mechanism to identify unknown objects. By integrating the results of close-set semantic segmentation and anomaly detection, we achieve effective feature-driven LiDAR open-set semantic segmentation.
Evaluations on both SemanticKITTI and nuScenes datasets demonstrate that our proposed framework significantly outperforms state-of-the-art methods. The source code will be made publicly available at \url{https://github.com/nubot-nudt/DOSS}.
\end{abstract}

\section{Introduction}
\label{sec:intro}
Semantic segmentation of LiDAR data is crucial for enabling autonomous robots to gain a high-level understanding of surrounding environments. Though it is challenging due to the sparsity of point cloud data, many existing works have demonstrated promising performance~\cite{milioto2019iros, langer2020iros, dewan2020icra}. However, most of these methods are limited to close-set semantic segmentation (CSS), where all classes to be segmented are known and labeled during training. In real open-world scenarios, it is impractical to label every possible class, making it important for autonomous robots to also detect unknown objects. This task, known as open-set semantic segmentation (OSS), combines CSS for known classes with anomaly detection for unknown classes. OSS is critical for autonomous robotic tasks, such as avoiding unknown animals or obstacles on the road. However, OSS presents significant challenges, as it requires identifying points belonging to unknown classes without labels during training.

\begin{figure}[t]
	\centering
	\includegraphics[width=1.0\linewidth]{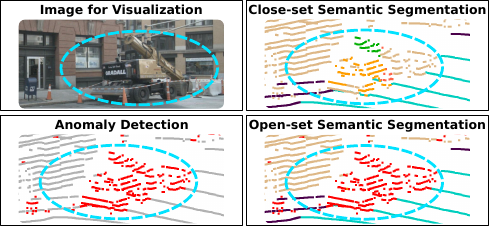}
	\caption{Visualization of open-set semantic segmentation. Close-set segmentation (CSS) (top right) only predicts the known classes while recognizing the unknown \textit{construction vehicle} in the blue ellipses as other known classes. Our method realizes anomaly detection (bottom left), \ie, segments unknown objects, and keeps the ability of CSS. Combining the two results above, we can finally achieve open-set semantic segmentation on LiDAR data (bottom right).}
	\label{fig:motivation}
\end{figure}

Several approaches have been proposed for LiDAR-based OSS. They either reorganize training data by excluding certain classes as unknowns~\cite{cen2022eccv}, or use adversarial networks to implicitly learn unknown class features~\cite{li2023cvpr}, which introduce artificial noise and are typically hard to train.
Another type~\cite{riz2023cvpr,deng2023iros} uses online clustering and uncertainty estimation to generate pseudo-labels by assuming a prior knowledge of unknown class numbers. Relying on such assumptions, these methods may underperform in real-world scenarios.

In this paper, we propose an effective feature-oriented framework to tackle the OSS task on LiDAR data. As illustrated in~\figref{fig:motivation}, we decompose the OSS task into CSS for known classes and anomaly detection for unknown classes. We propose a dual-decoder neural network that simultaneously achieves CSS and extracts distinct features suitable for realizing anomaly detection. A multi-objective loss function is carefully designed to guide the network in mapping known class features onto the surface of a hypersphere in feature space while clustering unknown class features at the center. Using these distinct features, our method estimates the confidence score for unknown objects based on maximum logits, enabling accurate anomaly detection. Leveraging the decomposed framework, our method achieves superior performance in detecting unknown classes while preserving strong CSS performance. We thoroughly evaluate our method on both SemanticKITTI and nuScenes datasets, and the experimental results validate that our method exceeds the state-of-the-art baseline method in LiDAR OSS.

In summary, the contributions of our work are threefold:
\begin{itemize}
    \item We propose an effective decomposed framework, achieving state-of-the-art performance in LiDAR OSS.
    \item We propose a dual-decoder network structure to maintain good CSS performance while extracting distinct features for unknown object anomaly detection.
    \item We propose a multi-objective loss function to promote the clear separation of the features of known and unknown objects in the high-dimensional feature space.
\end{itemize}

\section{Related Work}
\label{sec:related}

\textbf{3D Close-set Semantic Segmentation:}
3D CSS aims to estimate the class of each point, where the predicted classes are all known during training. According to the forms of point-cloud input for the backbones of deep learning, the 3D CSS methods can be mainly divided into point-based, range image-based, and voxel-based methods. 
PointNet~\cite{qi2017cvpr} is the first approach that directly works on the raw point cloud, which utilizes shared multi-layer perceptron to extract point-wise features and max pooling to obtain global features. PointNet++~\cite{qi2017nips} introduces a hierarchical neural network based on PointNet to learn local features with increasing contextual scales. After that, many point-based methods~\cite{hua2018cvpr,wu2019cvpr,huang2018cvpr} have been proposed to handle the 3D semantic segmentation problem. Range image-based methods~\cite{milioto2019iros,wu2018icra-scnn,ando2023cvpr} project the point cloud to the spherical range image. With the range image as the input, these methods leverage 2D convolutional neural networks to extract point-cloud features. Voxel-based methods~\cite{li2017iros,rethage2018eccv,zhu2021cvpr} project each point into the corresponding 3D structured voxels based on the point-cloud coordinate, so as to obtain the regular voxel input. The voxel-based network usually predicts the semantic label for each voxel and finally projects the labels back to each point. Some voxel-based methods utilize the fixed voxel size to organize the point cloud. However, the voxel size must be fully considered for the coverage of the point cloud and the computational cost. 
These semantic segmentation methods work well under the 3D close-set assumption but can not handle the open-set problem.

\textbf{2D Open-set Semantic Segmentation:}
Recently, 2D open-set semantic segmentation has gained significant attention due to its practical relevance in real-world applications. Maximum softmax probabilities (MSP)~\cite{hendrycks2017iclr} or maximum logit (MaxLogit)~\cite{hendrycks2022icml} are the commonly used baselines for predicting the uncertainty of unknown objects. Some works like MC-Dropout~\cite{gal2016icml}, Ensembles~\cite{lakshminarayanan2017nips}, and Sapkota~\etalcite{sapkota2022cvpr} realize open-set semantic segmentation by using Bayesian learning. The other trend of 2D open-set semantic segmentation is using generative models. These works~\cite{lis2019iccv,kong2021iccv,zhao2023cvpr} use generative adversarial networks (GAN) to generate and discriminate the objects of known classes, as well as the differences between the known and unknown ones. Recent work by Sodano~\etalcite{sodano2024cvpr} predicts the uncertainty of each pixel by the corresponding features, thus distinguishing the known and unknown objects in the feature space.

\textbf{3D Open-set Semantic Segmentation:}
Although open-set semantic segmentation methods have been explored for 2D vision, few works address the 3D OSS task. Some efforts~\cite{boudjoghra2023nips, xu2024cvpr} focus on dense point clouds generated by RGB-D sensors in indoor environments, but these methods often perform suboptimally when applied to the sparser LiDAR point clouds.
Recently, Cen~\etalcite{cen2022eccv} utilize Cylinder3D~\cite{Cylinder3D} as the backbone, resizing the known objects to synthesize pseudo unknown objects for training. However, the resized objects disorganize the point cloud in a way that influences the final results. Li~\etalcite{li2023cvpr} propose an open-set semantic segmentation method called Adversarial Prototype Framework (APF), which contains a prototypical constraint module for estimating the features of the known class and a feature adversarial module for discriminating the features not belonging to the known ones. This method utilizes GAN to estimate the features of unknown objects implicitly but also brings in the uncertainty of training. Riz~\etalcite{riz2023cvpr} acquire the unknown points by pseudo-labeling the points of novel classes with the exploited uncertainty quantification. However, the number of novel classes has been foregone and are used in network settings.

\section{Our Approach}
\label{sec:main}
\begin{figure*}[th]
	\centering
	\includegraphics[width=1\textwidth]{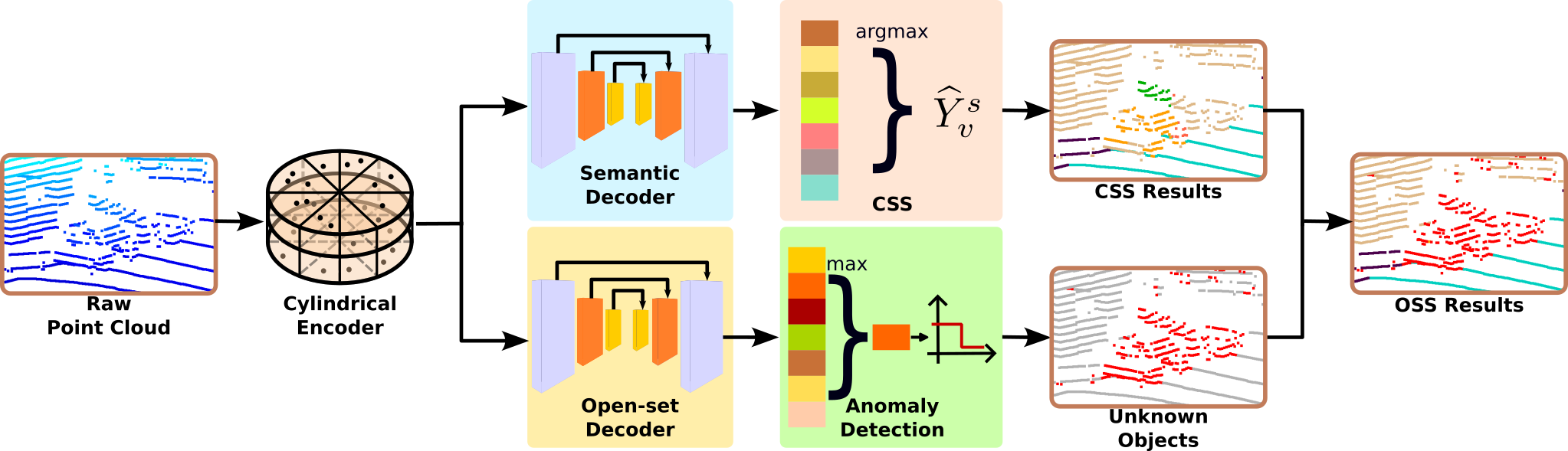}	
	\caption{Framework Overview: We first project points to the cylindrical voxels and extract the point-wise feature from the raw point cloud in the cylindrical encoder, so as to obtain aggregated voxel-wise features. These voxel features are fed to the dual decoders, \ie, semantic decoder and open-set decoder, generating distinct voxel features for guiding the known classes CSS and the anomaly detection of unknown objects. The close-set semantic results and the detected unknown objects are finally combined to realize effective open-set segmentation.}
	\label{fig:overview}
\end{figure*}

In this work, we propose a novel decomposed feature-oriented framework to tackle the LiDAR-based OSS task, named DOSS. The overview of our framework is shown in~\figref{fig:overview}, which contains five components: cylindrical encoder, semantic decoder, close-set semantic segmentation, open-set decoder, and anomaly detection. The cylindrical encoder projects points into cylindrical voxels, using the raw point cloud to extract voxel-wise features. Each voxel feature is represented by the largest point-wise feature within that voxel. The semantic decoder and open-set decoder further process these voxel features to generate distinct outputs for different tasks. The semantic decoder produces voxel features for CSS, predicting known classes for each voxel as a classification task. The open-set decoder distinguishes between known and unknown classes, aiding anomaly detection by identifying voxels with unknown objects based on the maximum logit of the learned voxel features. By replacing the close-set labels of detected anomalies with unknown flags, our framework achieves effective open-set semantic segmentation. We regard the labels of each point in one voxel as the same ones, obtaining the final point-wise OSS. Details of each component are discussed below. 

\subsection{Network Structure}
\label{sec:network}
Our network employs Cylinder3D~\cite{Cylinder3D} as the backbone to extract distinctive voxel features from LiDAR data, given its strong performance in LiDAR segmentation demonstrated in prior works~\cite{yang2023icra, wang2024segnet4d}. Cylinder3D projects point clouds into cylindrical voxels and utilizes a cylindrical encoder to extract voxel-wise features. We build upon this by retaining the cylindrical encoder and introducing two asymmetrical 3D convolutional networks as separate decoders for the close-set and open-set tasks. Both decoders share the same structure, containing upsampling and downsampling processes with asymmetrical sparse convolutions implemented. 

The output voxel feature dimension of the final logits layer in the close-set decoder is set to the number of known classes $K_s$. Using these features, the semantic class for each voxel is determined by selecting the highest-valued class. This process can be formulated as follows:
\begin{equation}
	\label{equ:close-set}
	\begin{split}
		\widehat{Y}^s_v = \argmax(\v{f}^s_v),
	\end{split}
\end{equation}
where the $\v{f}^s_v \in \RR^{1 \times K_s}$ and $\widehat{Y}^s_v \in \{1, \dots, K_s\}$ represent the features and the predicted known class of the voxel $v$.

The open-set decoder, which captures feature differences between known and unknown objects, generates voxel-wise features $\v{f}^o \in \RR^{n_v \times K_s}$, where $n_v$ denotes the number of voxels. These voxel-wise features are subsequently utilized for anomaly detection to achieve open-set segmentation, detailed in~\secref{sec:ab-detect}.

Our dual-decoder structure provides two key advantages for the OSS task: it minimizes mutual interference between the tasks by allowing each decoder to independently extract voxel features, and it enhances the network's learning capacity by decomposing the tasks, each supervised by distinct objective loss functions. The closed-set decoder is designed for the standard CSS task, while the open-set decoder focuses on voxel feature extraction and anomaly detection. Using task-specific decoders, we preserve the effectiveness of closed-set segmentation while enabling open-set semantic segmentation. The effectiveness of this structure is demonstrated in the ablation study in~\secref{sec:ab-study}.

\subsection{Anomaly Detection}
\label{sec:ab-detect}
Based on the voxel features predicted by the open-set decoder, we perform anomaly detection to segment unknown objects. Our multi-objective loss function encourages the network to cluster the voxel features of unknown objects near the center of the high-dimensional feature space, while ensuring those of known objects remain on the hypersphere's surface. The closer a voxel feature is to the center, the smaller its value. Using this design, we distinguish between known classes and unknown objects by analyzing the maximum values of the corresponding voxel features. We search for the highest value of each voxel feature, as it implicitly indicates the likelihood of the voxel belonging to a known or unknown object. Voxels with larger maximum feature values, lying on the hypersphere's surface, are more likely to belong to known objects, while those with smaller values are more likely to represent unknown objects. Therefore, during inference, if the maximum voxel feature is lower than a certain threshold, we consider this voxel as a part of unknown objects, \ie, anomaly. Mathematically, we look for the maximum value $\mathcal{M}_v$ of voxel feature $\v{f}^o_v$:
\begin{equation}
\label{equ:max-value}
	\begin{split}
	\mathcal{M}_v = \max(\v{f}^o_v).
	\end{split}
\end{equation}
Then, we utilize the maximum value to determine the open-set confidence score $\mathcal{S}_v$:
\begin{equation}
\label{equ:mls}
	\begin{split}
		\mathcal{S}_v = \left\{\begin{array}{cc}1.0&\text{, if }\mathcal{M}_v \leq \xi\\ 0.1 &\text{, if }\mathcal{M}_v > \xi\end{array}\right.,
	\end{split}
\end{equation}
where $\xi$ is the threshold parameter. We classify a voxel as unknown when its confidence score $\mathcal{S}_v$ equals 1. 

This approach allows us to leverage the designed voxel features to identify unknown object voxels in a feature-oriented manner. The labels of anomaly-detected unknown voxels in the CSS results are then marked as \textit{unknown}, while other voxels retain their original CSS labels. The updated labels, which incorporate both known CSS results and unknown anomaly detection outcomes, form the final OSS result obtained by our method.

\subsection{Multi-Objective Loss Function}
\label{sec:loss}
To train our dual-branch network for simultaneously achieving CSS and feature extraction for anomaly detection, we propose a set of multi-objective loss functions. These losses optimize the mean intersection over union (mIoU) for CSS and generate distinct voxel features to differentiate known from unknown classes for anomaly detection. Specifically, we employ five loss functions:  the weighted cross-entropy loss $\mathcal{L}_\mathrm{CE}$ and the lovasz-softmax loss $\mathcal{L}_\mathrm{LS}$ for the close-set decoder to optimize the close-set mIOU, together with the object-sphere loss $\mathcal{L}_{\mathrm{obj}}$, the contrastive loss $\mathcal{L}_{\mathrm{cont}}$, and the center loss $\mathcal{L}_\mathrm{cent}$ for the open-set decoder to learn distinct voxel features between known and unknown classes. 

For the CSS task, we follow Hu~\etalcite{hu2020cvpr}, implementing the weighted cross-entropy loss $\mathcal{L}_\mathrm{CE}$ and lovasz-softmax loss $\mathcal{L}_\mathrm{LS}$~\cite{berman2018cvpr}, to make the predicted probability distribution from the network as close as possible to the true probability distribution of known classes and optimize the mIOU:
\begin{equation}
\label{equ:ce-loss}
	\begin{split}
		\mathcal{L}_\mathrm{CE} = CE(\widehat{\v{Y}}^s, \v{Y}^s),
	\end{split}
\end{equation}
\begin{equation}
\label{equ:ls-loss}
	\begin{split}
		\mathcal{L}_\mathrm{LS} = LS(\widehat{\v{Y}}^s, \v{Y}^s),
	\end{split}
\end{equation}
where the $CE$, $LS$ are weighted cross-entropy and lovasz-softmax.
$\widehat{\v{Y}}^s \in \RR^{n_v \times 1}$ and $\v{Y}^s \in \RR^{n_v \times 1}$ are the predicted labels from the close-set decoder and the ground truth of known classes for each voxel.
The labels of unknown voxels in $\v{Y}^s$ are ignored when training.

For the open-set decoder, we aim to guide the output voxel features of known objects toward the surface of a hypersphere in feature space, enhancing inter-class separation and minimizing intra-class variation. Meanwhile, we cluster the features of unknown voxels near the center of the feature space to enable efficient feature-based anomaly detection. To achieve this, we employ the object-sphere loss~\cite{dhamija2018nips} and contrastive loss~\cite{chen2020icml}, ensuring that voxel features of unknown objects are drawn to the center of the high-dimensional space while voxel features of the same known class remain more consistent. The object-sphere loss is defined as follows:
\begin{equation}
\label{equ:obj-loss}
	\begin{split}
		\left.\mathcal{L}_{\mathrm{obj}}=\left\{\begin{array}{cc}\max\left(\eta-\|\v{f}^o_v\|^2,0\right)&\text{, if }v\in\bV^s\\\|\v{f}^o_v\|^2&\text{, otherwise}\end{array}\right.\right.,
	\end{split}
\end{equation}
where $\eta$ and $\bV^s$ represent the radius of the hypersphere and the set of voxels of known objects, respectively.

The contrastive loss ensures that voxel features from the open-set decoder are more similar within the same class while remaining distinguishable from voxel features of other classes:
\begin{equation}
\label{equ:cont-loss}
	\begin{split}
		\mathcal{L}_{\mathrm{cont}} = -\sum_{k=1}^{K_s}\log \frac{\exp{({\overline{\boldsymbol{f}}^{oF}_k}^\top {\overline{\boldsymbol{\mu}}^o_k}/\tau)}} {\sum_{i=1}^{K_s} \exp{({\overline{\boldsymbol{f}}^{oF}_k}^\top {\overline{\boldsymbol{\mu}}^o_i}/\tau)}},
	\end{split}
\end{equation}
where ${\overline{\boldsymbol{f}}^{oF}_k}$ is the mean voxel feature of known class $k$ calculated in the current point-cloud frame, and ${\overline{\boldsymbol{\mu}}^o_k}$ represents the mean voxel feature of class $k$ from the last epoch.
$\tau$ is the temperature parameter for contrastive learning.

\begin{figure}[t]
	\centering
    \subfloat[Expected feature distr.]{
			\includegraphics[width=3.0cm]{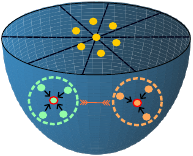}
			\label{fig:expected_features}
		}%
    \subfloat[Actual distribution visualized by tSNE.]{
        \includegraphics[width=4.8cm]{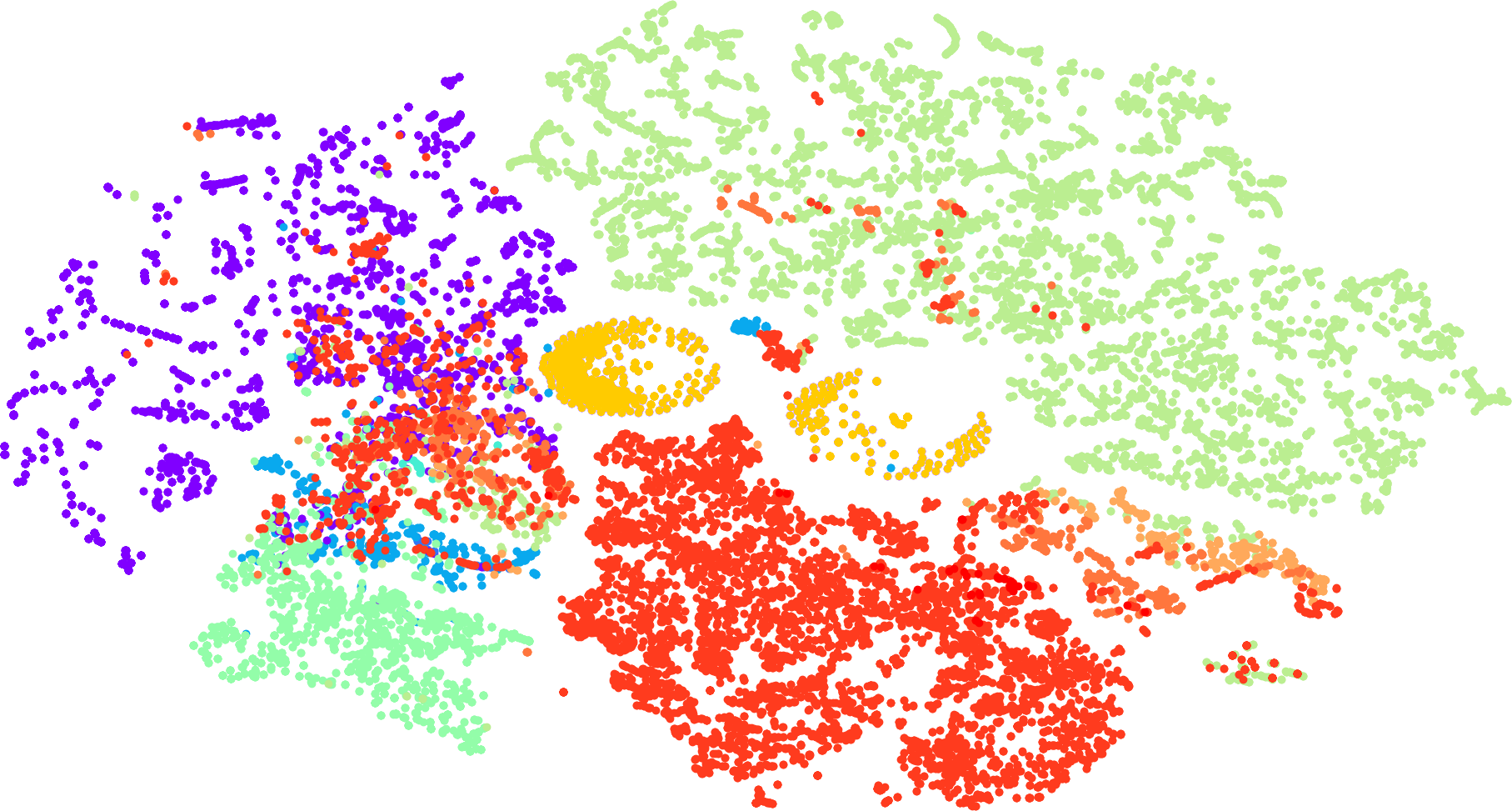}
        \label{fig:tsne_features}
        }
	\caption{Voxel features distribution generated by our designed open-set decoder. (a) The expected feature distribution: The faint blue region represents the surface of the hypersphere in high-dimensional feature space. Points on the hypersphere's surface with different colors correspond to different known classes. The red-circled points indicate the mean features of these classes. Yellow points at the center of the hypersphere represent features of unknown objects. The object-sphere loss clusters known class features on the hypersphere's surface and pushes unknown object features toward the center, creating a distinct separation between known and unknown features. The contrastive loss and center loss further cluster and tighten features of the same known class while separating them from different ones, reinforcing the gap between known classes and unknown objects. (b) The tSNE visualization of the actual feature distribution generated by our open-set decoder with one LiDAR frame. Each color indicates one class. Unknown class features (yellow points) are gathered in the center and far from that of known classes. The gaps between known classes are also obvious.}
	\label{fig:features}
\end{figure}

We additionally set a center loss for the open-set decoder, so as to tighten the distribution of voxel features for each class and to reduce the intersection between the voxel features of known and unknown objects:
\begin{equation}
\label{equ:cent-loss}
	\begin{split}
		\mathcal{L}_\mathrm{cent} = \sum_{k=1}^{K_s} \sum_{i=1}^{N_k} \|\v{f}^o_{ik} - \overline{\v{f}}^{oE}_k\|^2,
	\end{split}
\end{equation}
where ${N_k}$ is the voxel number of class $k$, $\v{f}^o_{ik}$ is the feature of the $i$-th voxel belonging to class $k$, and $\overline{\v{f}}^{oE}_k$ is the average voxel feature of previously encountered voxels in class $k$ updated in every training batch. The expected distribution of voxel features $\v{f}^o$ from the open-set decoder is shown in~\figref{fig:expected_features}, where the voxel features of known and unknown classes are on the surface of hypersphere and in the center of the feature space, respectively. For the known classes, the inter-class distance is increased and the inner-class distance is reduced. We also visualize the actual voxel feature distribution by using the feature visualization tool tSNE, see~\figref{fig:tsne_features}. The results show that our method can push the voxel features of the unknown classes to the center of the feature space and produce distinct feature gaps between known and unknown classes. The differences between known classes are also learned by our method.   

The loss functions mentioned above constitute our final loss for training:
\begin{equation}
\label{equ:final-loss}
	\begin{split}
		\mathcal{L} = \lambda_1 \mathcal{L}_\mathrm{CE} + \lambda_2 \mathcal{L}_\mathrm{LS} + \lambda_3 \mathcal{L}_{\mathrm{obj}} + \lambda_4 \mathcal{L}_{\mathrm{cont}} + \lambda_5 \mathcal{L}_{\mathrm{cent}},
	\end{split}
\end{equation}
where $\lambda_1, \lambda_2, \lambda_3, \lambda_4, \lambda_5$ are the hyper-parameters to adjust the weight of each loss function. Due to the proposed effective dual-decoder network structure for handling CSS and anomaly detection missions with respective decoders, our proposed loss setting can work without mutual interference between each loss function.

\section{Experimental Evaluation}
\label{sec:exp}

The main focus of this work is a feature-oriented framework for open-set semantic segmentation on LiDAR data.
We present our experiments to show the capabilities of our method.
The results of our experiments also support our key claims, which are:
(i)~our method achieves the state-of-the-art open-set semantic segmentation performance on LiDAR data;
(ii)~our proposed dual-decoder network structure can extract distinct voxel features for anomaly detection, while keeping the performance of CSS;
(iii)~our proposed loss settings can promote the voxel features of known and unknown objects to be separated, so as to realize effective open-set semantic segmentation.

\subsection{Experimental Setup}
\textbf{Dataset:}
Following the setup and the assumption of Cen~\etalcite{cen2022eccv}, we evaluate the performance of our method in the nuScenes dataset~\cite{caesar2020cvpr} and the SemanticKITTI dataset~\cite{behley2021ijrr}. The unknown classes used for evaluation are \{\textit{barrier, construction-vehicle, traffic-cone, trailer}\} in nuScenes and \{\textit{other-vehicle}\} in SemanticKITTI, respectively. The data from these classes is not used for network training. Besides, the data splits for training and evaluating are also the same as that of Cen~\etalcite{cen2022eccv}.

\textbf{Evaluation Metrics: }
We use the same open-set semantic segmentation evaluation metrics as the previous work~\cite{cen2022eccv} do:
mean intersection over union (mIoU) for the closed-set semantic segmentation, area under the precision-recall curve (AUPR), and area under the receiver operating characteristic curve (AUROC) for the open-set evaluation. The higher the three scores, the better the method for performing open-set semantic segmentation.

\textbf{Implementation Details: }
We follow Cylinder3D~\cite{Cylinder3D} to use Adam as the optimizer and set the learning rate to $1e^{-3}$.
For nuScenes dataset, we set $\eta = 1.0$ in \eqref{equ:obj-loss}, $\tau = 0.1$ in \eqref{equ:cont-loss}, $\lambda_1 = 1.0$, $\lambda_2 = 1.0$, $\lambda_3 = 0.5$, $\lambda_4 = 0.5$, $\lambda_5 = 0.3$ in \eqref{equ:final-loss}, and $\xi = 0.65$ in \eqref{equ:mls}. For SemanticKITTI dataset, $\eta = 2.0$, $\lambda_3 = 0.9$, and $\xi = 0.4$. More details can be found in our open-source code.

\textbf{Compared Baselines: }
We compare our method with the state-of-the-art 3D point-cloud open-set methods REAL~\cite{cen2022eccv}, APF~\cite{li2023cvpr}. Since the code of APF is not available, we only compare it with the reported results on SemanticKITTI. We also use the implemented 2D anomaly detection method MSP~\cite{hendrycks2017iclr}, MaxLogit~\cite{hendrycks2022icml}, and MC-Dropout~\cite{gal2016icml} to combine with the CSS method Cylinder3D~\cite{Cylinder3D}, treating them as our baselines. Additionally, we import the 2D feature-based method ContMAV~\cite{sodano2024cvpr} to tackle the 3D task by using the projected range image of LiDAR data as the input. Since the usage of $\mathcal{L}_\mathrm{feat}$ in ContMAV destabilizes the training results, we remove this loss function when training and inferencing.




\subsection{Performance}


The first experiment evaluates the OSS performance of our decomposed-OSS approach DOSS in the nuScenes and SemanticKITTI dataset. 
The results presented in \tabref{tab:comparison-nusc} and \tabref{tab:comparison-semantickitti}
demonstrate that our method achieves state-of-the-art OSS performance on LiDAR data. Notably, in addition to the significant improvement in AUROC, our approach markedly increases the AUPR on the nuScenes and SemanticKITTI datasets while maintaining high CSS accuracy. The results of ContMAV further highlight the limitations of applying a 2D neural network to the OSS problem. Since methods like MSP~\cite{hendrycks2017iclr}, MaxLogit~\cite{hendrycks2022icml}, and MC-Dropout~\cite{gal2016icml} rely solely on CSS from Cylinder3D~\cite{Cylinder3D}, their anomaly detection performance is constrained. Approaches that address anomaly detection train networks to learn unknown object characteristics, which may negatively impact closed-set performance and reduce mIoU. In contrast, our method maintains competitive CSS performance while achieving the best OSS results.

\begin{figure*}[th]
		\centering
        \subfloat[Images for visualization]{
			\includegraphics[width=3.3cm]{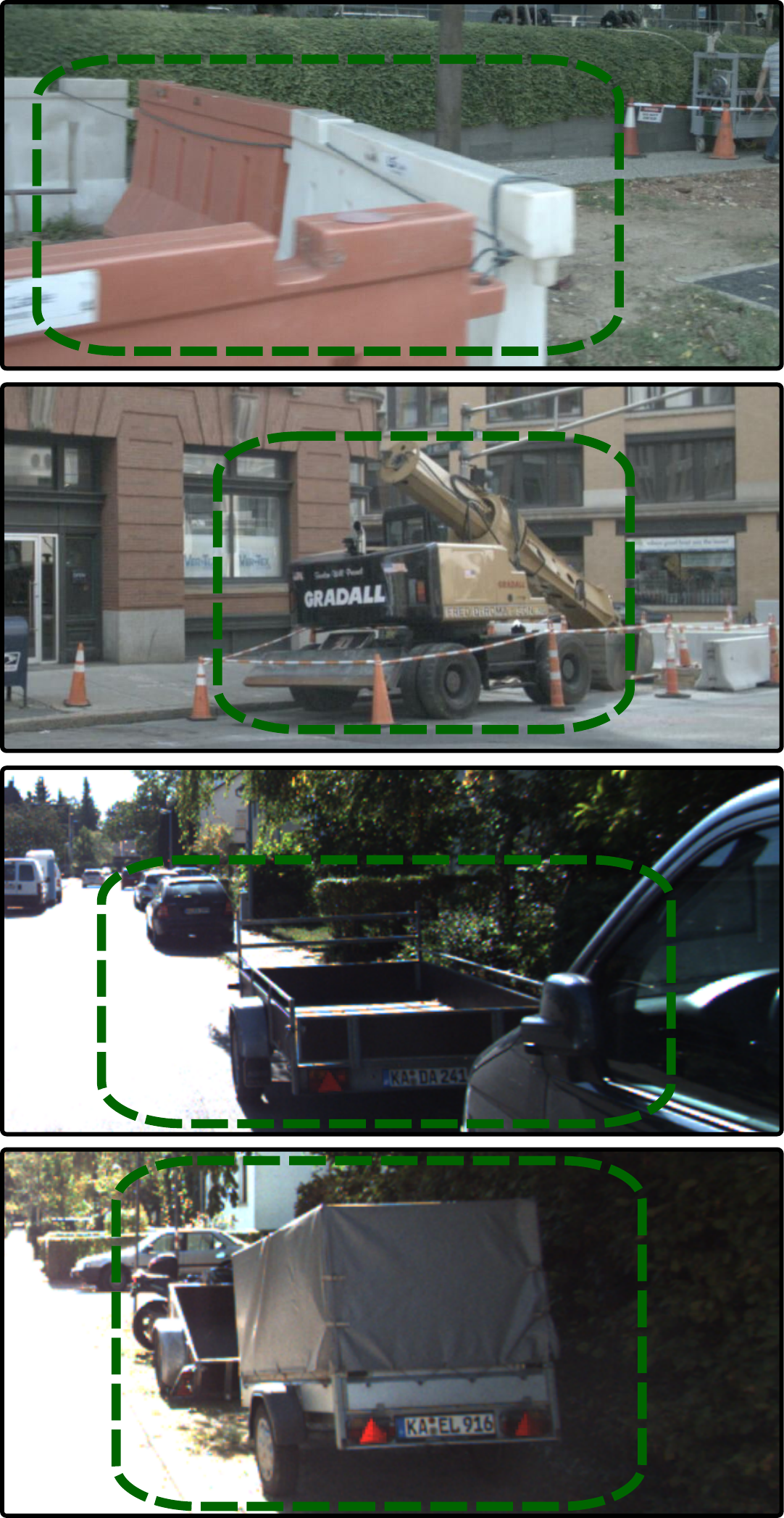}
			\label{fig:compare-RGB}
		}%
		\subfloat[REAL~\cite{cen2022eccv}]{
			\includegraphics[width=3.3cm]{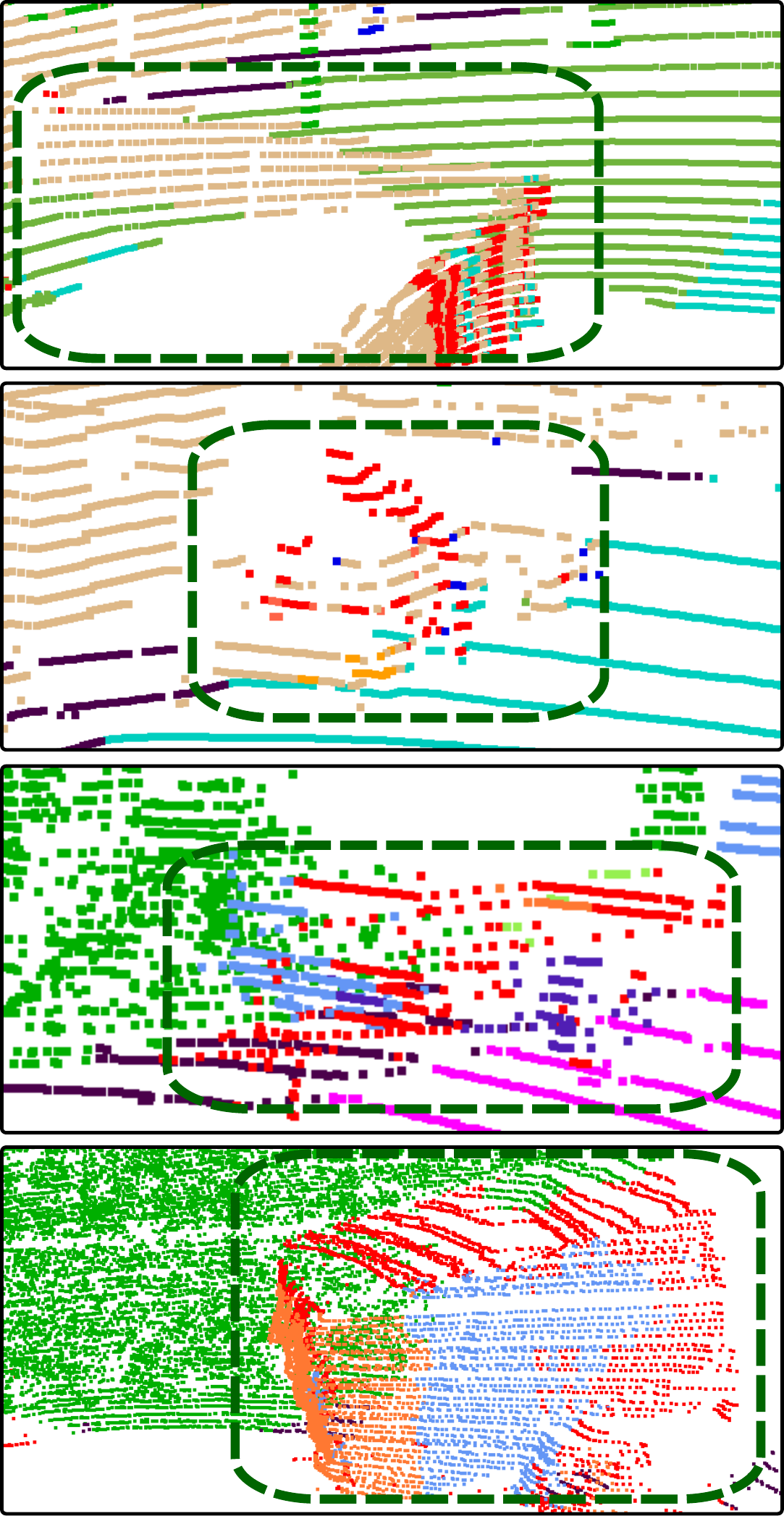}
			\label{fig:compare-REAL}
		}%
		\subfloat[ContMAV~\cite{sodano2024cvpr}]{
			\includegraphics[width=3.3cm]{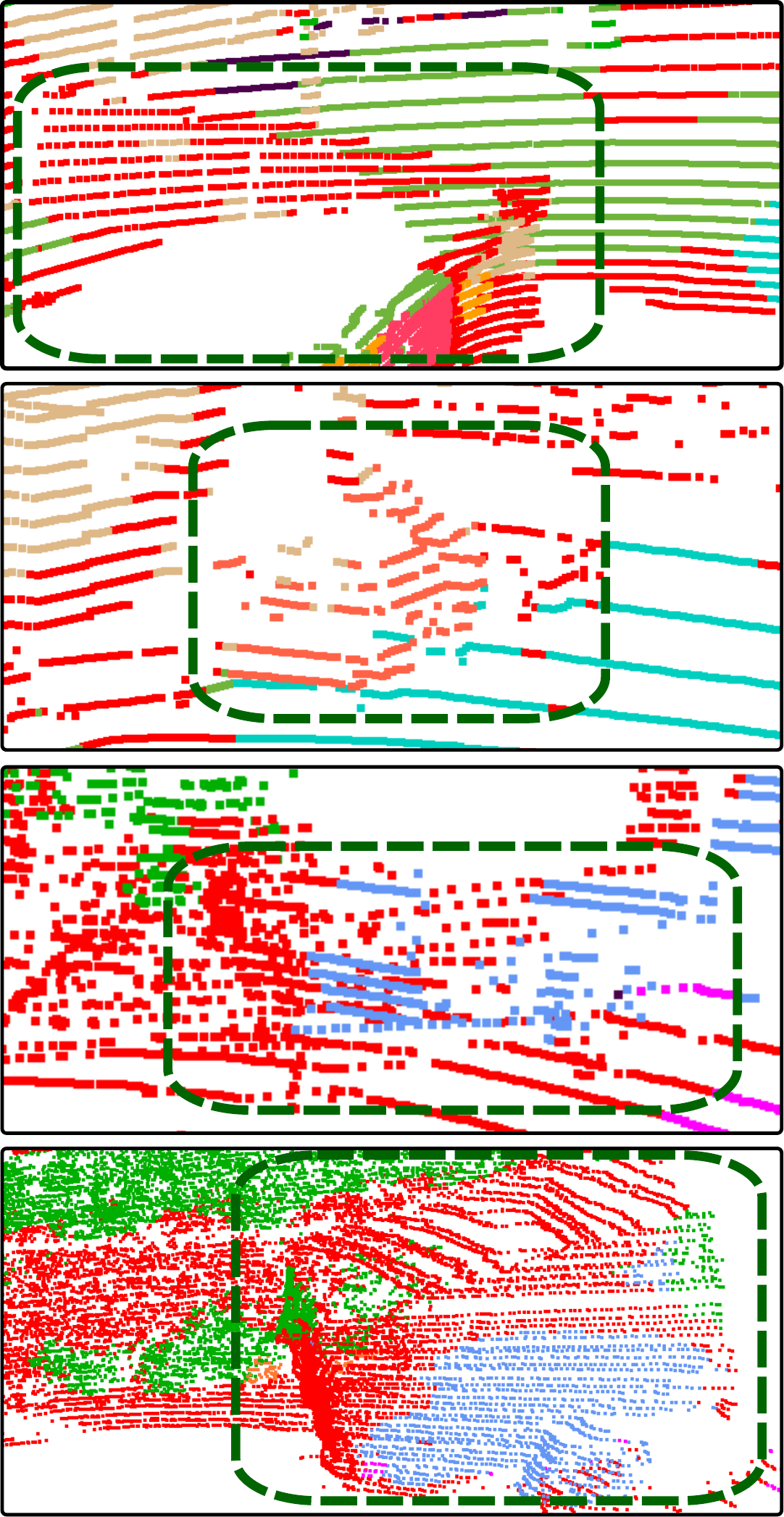}
			\label{fig:compareCONTMAV}
		}%
		\subfloat[DOSS (ours)]{
			\includegraphics[width=3.3cm]{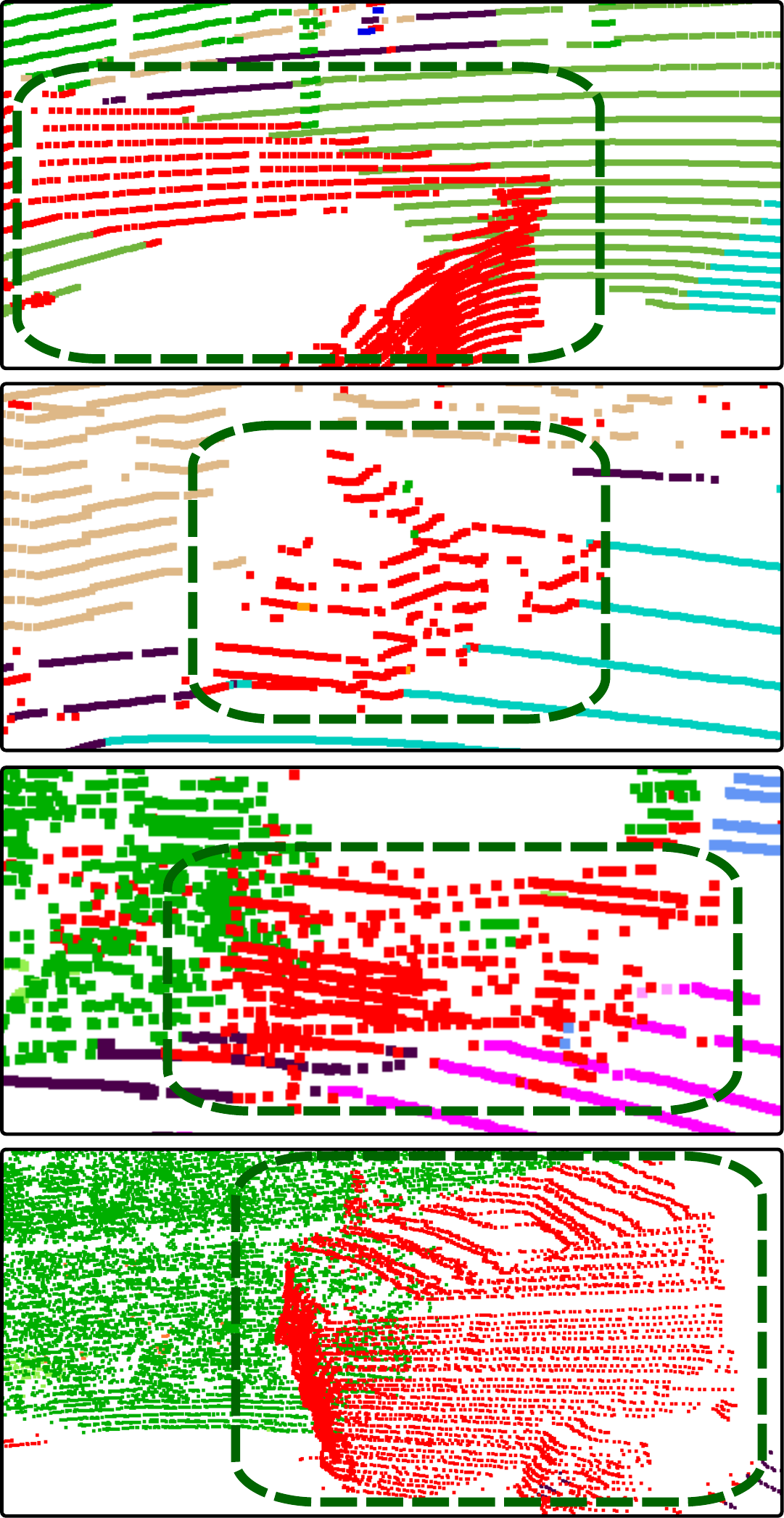}
			\label{fig:compare-DOSS}
		}%
		\subfloat[Ground truth]{
			\includegraphics[width=3.3cm]{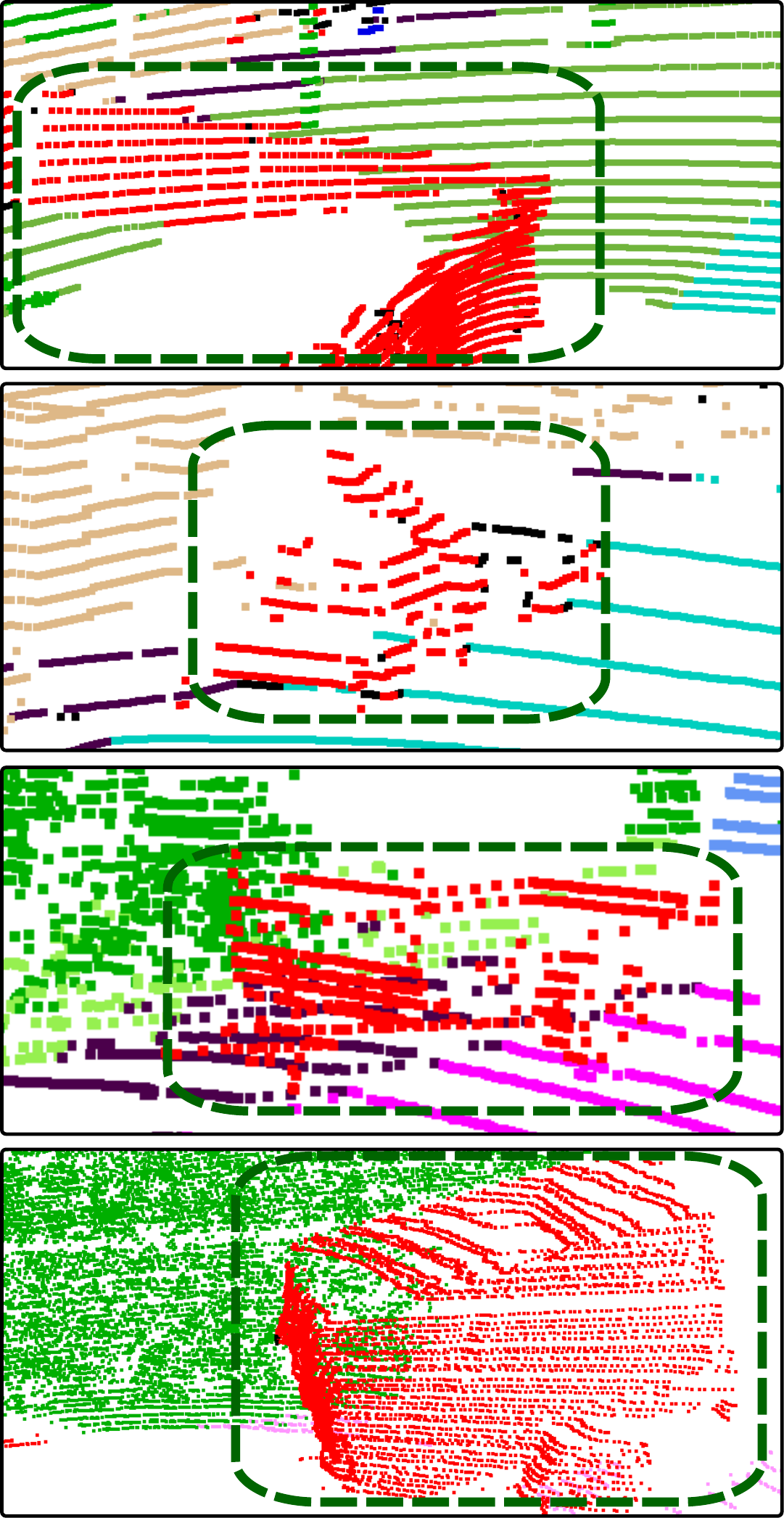}
			\label{fig:compare-GT}
		}%
		\caption{Visualization of OSS results in nuScenes dataset (first two rows) and SemanticKITTI (the last two rows). Red points belong to the unknown objects (circled with dark green dotted lines), \ie, \textit{barrier} in the first row, \textit{construction vehicle} in the second row, and \textit{other-vehicle} in the last two rows in these cases. As shown in the figures, REAL and ContMAV can not detect all unknown points. Especially, ContMAV regards many points of known classes as unknown ones. Our method manages to segment the complete unknown object with few under-segmentation.}
		\label{fig:qualitive-result}
		\vspace{-0.1cm}
\end{figure*}

\figref{fig:qualitive-result} visualizes the qualitative results of open-set semantic segmentation from our method compared to others. The visualization demonstrates that our approach effectively detects unknown object points while preserving accurate CSS, achieving accurate OSS.

For the runtime performance, our methods achieve 2.58 hz on a single NVIDIA GeForce RTX 3060 Laptop GPU with 64-beam LiDAR, while the runtime of 3D-LiDAR baseline REAL~\cite{cen2022eccv} is 2.77 hz. Our method achieves better OSS performance without sacrificing much runtime performance.

\begin{table}[t]
	\caption{Comparison of OSS performance using nuScenes dataset.}
	\centering
	
	\begin{tabular}{C{3cm}ccc}
		\toprule
		Methods & AUPR & AUROC & mIoU \\
		\midrule
		MSP  & 4.0 & 75.6 & 57.7 \\
        MaxLogit  & 7.9 & 79.0 & 57.7\\
        MC-Dropout  & 13.2 & 80.6 & 57.7\\
        \midrule
        ContMAV  & 28.0 & 59.4 & 42.7\\
		REAL  & 21.2 & 84.5 & 56.8\\
		\textbf{DOSS (Ours)}  & \textbf{51.8} & \textbf{89.9} & \textbf{57.9}\\
		\bottomrule
	\end{tabular}
    \label{tab:comparison-nusc}
\end{table}

\begin{table}[t]
	\caption{Comparison of OSS performance using SemanticKITTI.}
	\centering
	
	\begin{tabular}{C{3cm}ccc}
		\toprule
		Methods & AUPR & AUROC & mIoU \\
		\midrule
        MSP  & 5.2 & 72.2 & 57.0\\
        MaxLogit  & 5.0 &  68.9 & 57.0\\
        MC-Dropout  & 5.7 & 71.5 & 57.0\\
        \midrule
		ContMAV  & 24.6 & 62.0 & 33.4\\
		REAL   & 20.8 & 84.9 & \textbf{57.8}\\
        APF   & 36.1 & 85.6 & 57.3\\
		\textbf{DOSS (Ours)}  & \textbf{50.6} & \textbf{87.6} & 57.0\\
		\bottomrule
	\end{tabular}
    \label{tab:comparison-semantickitti}
\end{table}

\subsection{Ablation Study}
\label{sec:ab-study}
To validate the effectiveness of our proposed network structure and the loss settings, we conduct the following ablation studies to support our claims. Both of these two experiments are evaluated in the nuScenes dataset.

\textbf{Ablation study of the network structure: }
The first ablation study aims to prove that our proposed dual-decoder network structure can achieve promising anomaly detection results and maintain the performance of CSS. To this end, we set up two other network settings. Both of these networks use the same cylindrical encoder, while the first network, denoted as $\left[\mathrm{A}\right]$, only has one decoder with a one-layer output head. The second network, denoted as $\left[\mathrm{B}\right]$, also has only one decoder but is followed by two one-layer output heads for CSS and anomaly detection missions, respectively. Our proposed network, which contains dual decoders and the corresponding one-layer output heads for each decoder, is marked as $\left[\mathrm{C}\right]$. As for the loss settings, the single output head of network $\left[\mathrm{A}\right]$ utilizes the complete loss function $\mathcal{L}$. Both of network $\left[\mathrm{B}\right]$ and our proposed network $\left[\mathrm{C}\right]$ implement $\lambda_1 \mathcal{L}_\mathrm{CE} + \lambda_2 \mathcal{L}_\mathrm{LS}$ for the CSS head and $\lambda_3 \mathcal{L}_{\mathrm{obj}} + \lambda_4 \mathcal{L}_{\mathrm{cont}} + \lambda_5 \mathcal{L}_{\mathrm{cent}}$ for the other head.

The quantitative results of this experiment are shown in \tabref{tab:as_network}. Although network $\left[\mathrm{A}\right]$ achieves relatively higher AUPR and AUROC scores, probably because of the implicit use of the known class voxel features, the mIoU score for the CSS mission drops dramatically. Network $\left[\mathrm{B}\right]$ utilizes dual heads for each mission and obtains balanced results for open-set semantic segmentation. It proves that our idea of implementing dual heads for each mission in the OSS task is beneficial to realize effective anomaly detection and keep the comparative performance of CSS. However, network $\left[\mathrm{B}\right]$ only utilizes one decoder to obtain features for the whole OSS mission. It makes it hard for the network to adapt to the task with only a few parameters, thus leading to a suboptimal result. Our proposed network structure, \ie, network $\left[\mathrm{C}\right]$, leverages the idea of implementing dual heads for the OSS mission and dual decoder for more parameters. It acquires the highest mIoU score, comparable AUPR, AUROC scores relative to network $\left[\mathrm{A}\right]$, and higher AUROC scores than network $\left[\mathrm{B}\right]$. The experimental results validate our claims that our proposed network structure can extract distinct voxel features to achieve promising anomaly detection results, while keeping the CSS performance. 

\begin{table}[t]
	\caption{Ablation study on network structure.}
	\centering
	
	\begin{tabular}{C{2cm}ccc}
		\toprule
		  Networks & AUPR & AUROC & mIoU \\
		\midrule
		$\left[\mathrm{A}\right]$  & \textbf{52.6} & \textbf{91.0} & 29.9\\
		$\left[\mathrm{B}\right]$  & 51.8 & 87.9 & 57.3\\
		  $\left[\mathrm{C}\right]$  & 51.8 & 89.9 & \textbf{57.9}\\
		\bottomrule
  	\end{tabular}

  \label{tab:as_network}
\end{table}

\textbf{Ablation study of the loss settings: }
The ablation study on loss settings supports our third claim, demonstrating that the proposed loss configuration effectively separates voxel features of known and unknown objects, leading to improved open-set semantic segmentation. Our method utilizes five loss functions for network training, \ie, the weighted cross-entropy loss $\mathcal{L}_\mathrm{CE}$, the lovasz-softmax loss $\mathcal{L}_\mathrm{LS}$, the object-sphere loss $\mathcal{L}_\mathrm{obj}$, the contrastive loss $\mathcal{L}_\mathrm{cont}$, and the center loss $\mathcal{L}_\mathrm{cent}$. The weighted cross-entropy loss and the lovasz-softmax loss are used for the semantic decoder, while the object-sphere loss $\mathcal{L}_\mathrm{obj}$, the contrastive loss $\mathcal{L}_\mathrm{cont}$, and the center loss $\mathcal{L}_\mathrm{cent}$ are employed for the open-set decoder. In this experiment, we focus on the impact of the open-set loss functions, as the effectiveness of the close-set loss functions has been shown in prior work~\cite{Cylinder3D}.  
We add the object-sphere loss $\mathcal{L}_\mathrm{obj}$, the contrastive loss $\mathcal{L}_\mathrm{cont}$, and the center loss $\mathcal{L}_\mathrm{cent}$ in the network training step by step. 
As shown in~\tabref{tab:as_loss} the scores of AUPR and AUROC keep increasing as the loss functions are gradually added. Furthermore, the values of mIoU also increase with the addition of the loss functions to some extent. This experiment validates that every loss of our designed multi-objective function plays an important role in generating distinct voxel features of anomaly detection.

\begin{table}[t]
	\caption{Ablation study on loss settings.}
	\centering
	
	\begin{tabular}{cccccccc}
		\toprule
		$\mathcal{L}_\mathrm{CE}$ & $\mathcal{L}_\mathrm{LS}$ & $\mathcal{L}_\mathrm{obj}$ & $\mathcal{L}_\mathrm{cont}$ & $\mathcal{L}_\mathrm{cent}$
        & AUPR & AUROC & mIoU \\
		\midrule
		\checkmark & \checkmark & & &                                   & 17.0 & 27.7 & 46.6\\
		\checkmark & \checkmark & \checkmark & &                        & 49.7 & 52.1 & 43.0\\
		\checkmark & \checkmark & \checkmark & \checkmark &             & 50.7 & 60.2 & 57.8\\
		\checkmark & \checkmark & \checkmark & \checkmark &  \checkmark & \textbf{51.8} & \textbf{89.9} & \textbf{57.9} \\
		\bottomrule
	\end{tabular}
	
	\label{tab:as_loss}
\end{table}


\section{Conclusion}
\label{sec:conclusion}

In this paper, we introduce a novel decomposed feature-oriented framework for open-set semantic segmentation on LiDAR data. Our approach leverages a dual-decoder neural network and a multi-objective loss function to differentiate voxel features for both known and unknown classes. This enables accurate anomaly detection for unknown objects while maintaining robust close-set semantic segmentation performance. We implemented and evaluated our method on the SemanticKITTI and nuScenes datasets, comparing it with existing techniques. The results demonstrate that our method achieves state-of-the-art performance.



\bibliographystyle{ieeetr}

\bibliography{ms}

\end{document}

%% file: stachnisslab-latex.tex
\usepackage{graphics}           
\usepackage{times}              
\usepackage{amsmath}            
\usepackage{amssymb}            
\usepackage{graphicx}
\usepackage{algorithm}
\usepackage[noend]{algpseudocode}
\usepackage{booktabs}
\usepackage{color}
\definecolor{instructioncolor}{rgb}{.5,.5,.5}

\usepackage[font=small]{caption}

\def\secref#1{Sec.~\ref{#1}}
\def\figref#1{Fig.~\ref{#1}}
\def\tabref#1{Tab.~\ref{#1}}
\def\eqref#1{Eq.~(\ref{#1})}

\makeatletter
\usepackage{xspace}
\DeclareRobustCommand\onedot{\futurelet\@let@token\@onedot}
\def\@onedot{\ifx\@let@token.\else.\null\fi\xspace}
 
\def\ie{i.e\onedot}

\def\etal{{et al}\onedot}
\makeatother

\def\etalcite#1{\etal~\cite{#1}}

\usepackage{array}
\newcolumntype{L}[1]{>{\raggedright\let\newline\\\arraybackslash\hspace{0pt}}m{#1}}
\newcolumntype{C}[1]{>{\centering\let\newline\\\arraybackslash\hspace{0pt}}m{#1}}
\newcolumntype{R}[1]{>{\raggedleft\let\newline\\\arraybackslash\hspace{0pt}}m{#1}}

%% file: stachnisslab-math.tex
\def\argmax{\mathop{\rm argmax}}

\newcommand{\RR}{\mathbb{R}}

\renewcommand{\b}[1]{\mbox{\boldmath$#1$}}

\renewcommand{\v}[1]{{\b #1}} 

\newcommand{\bV}{\b V}